\documentclass[a4letter, 10 pt, conference]{ieeeconf}  

\IEEEoverridecommandlockouts                              

\usepackage{calrsfs}
\usepackage{graphicx}
\usepackage{gensymb}
\usepackage{footmisc}
\usepackage{booktabs}
\usepackage{multicol}

\usepackage{multirow}
\usepackage{longtable}
\usepackage{color,soul}
\usepackage{graphics} 
\usepackage{epsfig} 
\usepackage{times} 
\usepackage{amsmath} 
\usepackage{amssymb}  
\usepackage{mathtools}
\usepackage{caption}
\usepackage{subcaption}

\DeclareMathAlphabet{\pazocal}{OMS}{zplm}{m}{n}

\title{\LARGE \bf
DiSCO: Differentiable Scan Context with Orientation
}

\author{Xuecheng Xu, Huan Yin, Zexi Chen, Yue Wang and Rong Xiong 
\thanks{Xuecheng Xu, Huan Yin, Zexi Chen, Yue Wang and Rong Xiong are with the State Key Laboratory of Industrial Control Technology and Institute of Cyber-Systems and Control, Zhejiang University, Zhejiang, China. Yue Wang is the corresponding author {\tt\small wangyue@iipc.zju.edu.cn}.}%
}

\begin{document}

\maketitle
\thispagestyle{empty}
\pagestyle{empty}

\begin{abstract}

Global localization is essential for robot navigation, of which the first step is to retrieve a query from the map database. This problem is called place recognition. In recent years, LiDAR scan based place recognition has drawn attention as it is robust against the appearance change. In this paper, we propose a LiDAR-based place recognition method, named Differentiable Scan Context with Orientation (DiSCO), which simultaneously finds the scan at a similar place and estimates their relative orientation. The orientation can further be used as the initial value for the down-stream local optimal metric pose estimation, improving the pose estimation especially when a large orientation between the current scan and retrieved scan exists. Our key idea is to transform the feature into the frequency domain. We utilize the magnitude of the spectrum as the place signature, which is theoretically rotation-invariant. In addition, based on the differentiable phase correlation, we can efficiently estimate the global optimal relative orientation using the spectrum. With such structural constraints, the network can be learned in an end-to-end manner, and the backbone is fully shared by the two tasks, achieving interpretability and light weight. Finally, DiSCO is validated on three datasets with long-term outdoor conditions, showing better performance than the compared methods.
\footnote{\label{web}Codes are released at https://github.com/MaverickPeter/DiSCO-pytorch. Appendix is stored at the same repository.}
\end{abstract}


\section{Introduction}

Large-scale place recognition is an essential component of global localization in outdoor autonomous mobile robotics applications. It provides candidate places for loop-closure which further corrects the drift during the long-term running and builds drift-free globally consistent maps. Thanks to the rich information in vision sensors, lots of successful visual place recognition methods were presented \cite{Lowry2016,Arandjelovic2018,tang2020adversarial}. However, visual place recognition algorithms suffer from strong appearance and viewpoint changes, thus can not be robustly utilized in long-term outdoor scenarios. In contrast, 3D LiDAR captures more stable structural information and becomes a competitive sensor option for place recognition. In addition, based on the 360-degree view of the LiDAR sensor, LiDAR-based place recognition has a chance to recognize a scan captured at a similar place in the database that has a large orientation difference from the query scans. This is useful in practice, e.g., the database is built when running in one direction of a highway while the query scan is captured from the opposite direction.

However, many works on LiDAR-based place recognition directly imitate the visual place recognition pipeline ignoring the characteristics of LiDAR data \mbox{\cite{Steder2011, B2018}}. First, unlike vision, LiDAR data is represented as 3D point clouds. Hence network design useful for vision may be ineffective for LiDAR data. For instance, rotation within the image-plane is common in popular bird's-eye view (BEV) representation of LiDAR scan but rarely exists in an image of a front-facing camera, which makes difference for convolution because it is only effective for image-plane translation. Another challenge is that existing point cloud registration techniques, e.g., ICP \mbox{\cite{besl1992method}}, are very sensitive to the initialization. They usually fail when the initial rotation has a 180-degree flip even if the initial position is perfect, essentially leading to the failure of the loop closure. Thus, for successful loop closure, we have to unreasonably discard these recognized places with correct position but large orientation difference, or use an exhaustive search for the correct relative orientation.

\begin{figure}[t]
\centering
\includegraphics[width=0.47\textwidth]{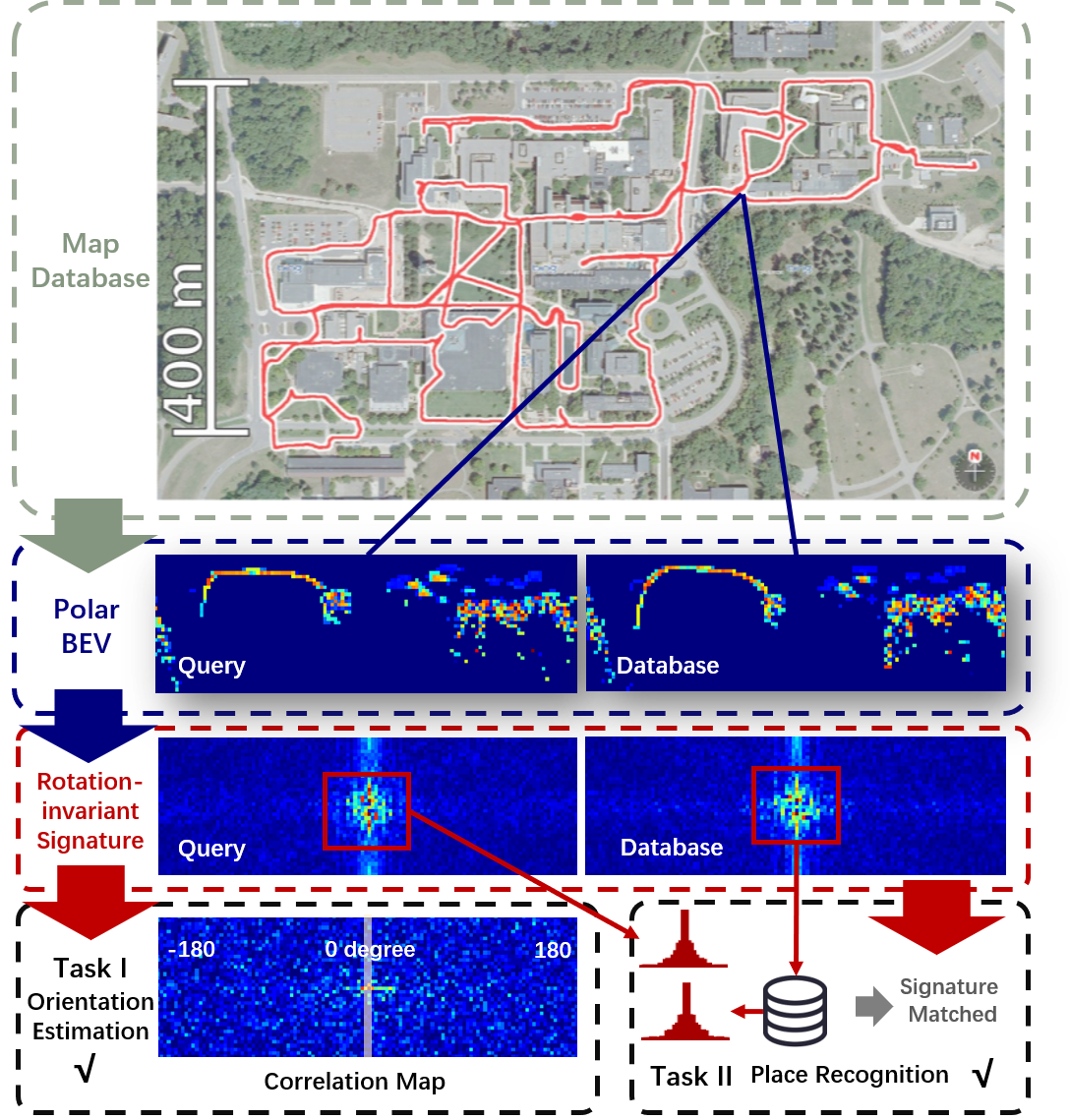}
\caption{\small{We propose DiSCO, a novel method for global localization, which jointly performs: place retrieval and relative orientation estimation. The lower side shows an example of similarity evaluation and estimating the relative orientation. The map image from \cite{carlevaris2016university} shows an overview of our database. By using the polar BEV and learned rotation-invariant signatures, a robot can localize to the map.}}
\label{fig:Teaser_img}
\vspace{-0.35cm}
\end{figure}

In this paper, we set to address the above issues by presenting a novel LiDAR-based place recognition method, called Differentiable Scan Context with Orientation (DiSCO) as shown in Fig. \ref{fig:Teaser_img}. Specifically, following the polar transformed bird's-eye view representation utilized in the Scan Context \cite{Kim2018}, our end-to-end learnable DiSCO has a highly explainable architecture to enforce the network to learn a rotation-invariant place signature. As rotation-invariance is converted into translation-invariance in the polar domain, we have a frequency spectrum based similarity evaluation metric that can be efficiently evaluated via Euclidean distance. In addition, for successful loop closure, we propose a differentiable phase correlation, which can reuse the frequency spectrum to generate a reliable relative orientation between the query and database scans. With this relative orientation as the initial value for scan registration, a precise relative pose can be expected. As the parameters are fully shared for both place retrieval and relative orientation estimation, our network architecture is lightweighted. To summarize, the contribution of the paper includes
\begin{itemize}
\item A highly interpretable network architecture is proposed to extract rotation-invariant place signature in the frequency domain, achieving efficient metric evaluation.
\item A differentiable phase correlation estimator is proposed for global optimal relative orientation estimation. This branch shares the network parameter with the signature extraction branch, further improving the efficiency.
\item Evaluations on three large scale datasets with multiple sessions and seasonal environmental changes demonstrates the effectiveness of DiSCO.
\end{itemize}

\section{Related Work}

A considerable amount of literature has been published on the place recognition task. These studies proposed different descriptors used to discriminatively represent places. Recent evidence suggests that 2D orientation can be estimated together with place candidates.

\subsection{Handcraft Descriptors}

Many works have been conducted for place recognition with 3D point clouds. Most of the previous works designed handcraft descriptors that describe either local or global features of a point cloud to discriminate places. Early approaches of 3D analysis using handcraft descriptors mainly adopt statistics ideas of histograms or signatures \cite{Tombari2011}. Signature-based methods \cite{Stein1992, Knopp2010} voxelize a point cloud into different regions and calculate interest values of the regions, and aggregate information from all regions. Histogram-based approaches \cite{Rusu2010} count the trait values of each point and generate the descriptor. These handcraft methods focused on acquiring local features and manually aggregate them as global descriptors which are not efficient for complex scenes. Other works followed the visual place recognition idea that extracting keypoints on point clouds by projecting the point cloud to a range image. Steder et al. \cite{Steder2011} presented a pipeline that uses NARF keypoint extracted on range images and generated global descriptors using bag-of-words. Following this projecting trend, He et al. \cite{He2016} proposed M2DP which projects point cloud to multiple planes as a global descriptor.

\subsection{Deep Learning Descriptors}

With the big progress made by deep learning in computer vision, learning-based approaches have been proposed recently. A common idea is to learn the embedding of places and matching between them. Barsan et al. \cite{B2018} presented a method that embeds point cloud with intensity information and achieved place retrieval by matching embeddings. Dube and Cramariuc \cite{Dube2017, cramariuc2018learning} preferred learning embeddings for segments. Similar to the handcraft descriptors, Yin et al. \cite{yin2018locnet,Yin2020} proposed a semi-handcrafted feature learning framework using the siamese network. Kim et al. \cite{Kim2019} also developed a semi-handcraft descriptor called Scan Context which transforms point clouds to polar images and uses simple CNN for long-term place classification. Other methods tried to learning descriptors directly from 3D points. PointNetVLAD \cite{Uy2018} is consist of point feature network PointNet \cite{Qi2017} and the visual place recognition method NetVLAD \cite{Arandjelovic2018} and generates global descriptor for 3D point clouds. However, these methods just consider the place retrieval part of global localization. PointNetVLAD discards the pose information at the very first steps. Scan Context also eliminated the rotation factor by augmenting the input. Eventually, many researchers realized this issue and some methods appear to solve the place recognition with yaw estimation. Schaupp et al. \cite{Schaupp2019} proposed OREOS that not only estimate place candidates but also estimated the yaw discrepancy between scans. OverlapNet \cite{Chen2020} estimated yaw with range images exploiting multiple types of information extracted from 3D LiDAR scans.

\subsection{Pose Estimation}

For pose estimation on the point cloud, traditional approaches followed the idea of ICP and suffer from the quality of the initial pose. Some learning-based methods' solution is to directly regress the pose but lacks interpretability. In our case, we mainly estimate the relative orientation between two BEVs. OREOS \cite{Schaupp2019} borrowed the idea of regressing the pose and achieved good performance. Barnes et al \cite{Barnes2019} deployed a correlation-based matching method to estimate translation but exhaustively search for the orientation. DPCN \cite{chen2020deep} modified the traditional correlation-based image registration pipeline to a differentiable manner and achieved 3-DoF pose estimation on heterogeneous sensor measurements. Recent works \mbox{\cite{bulow2018scale,bulow2020divide}} show the power of the Fourier-Mellin Transform in 3D registration tasks which is an extension of 2D transform adopted in our method.

\begin{figure*}[t]
\centering
\includegraphics[scale=0.5]{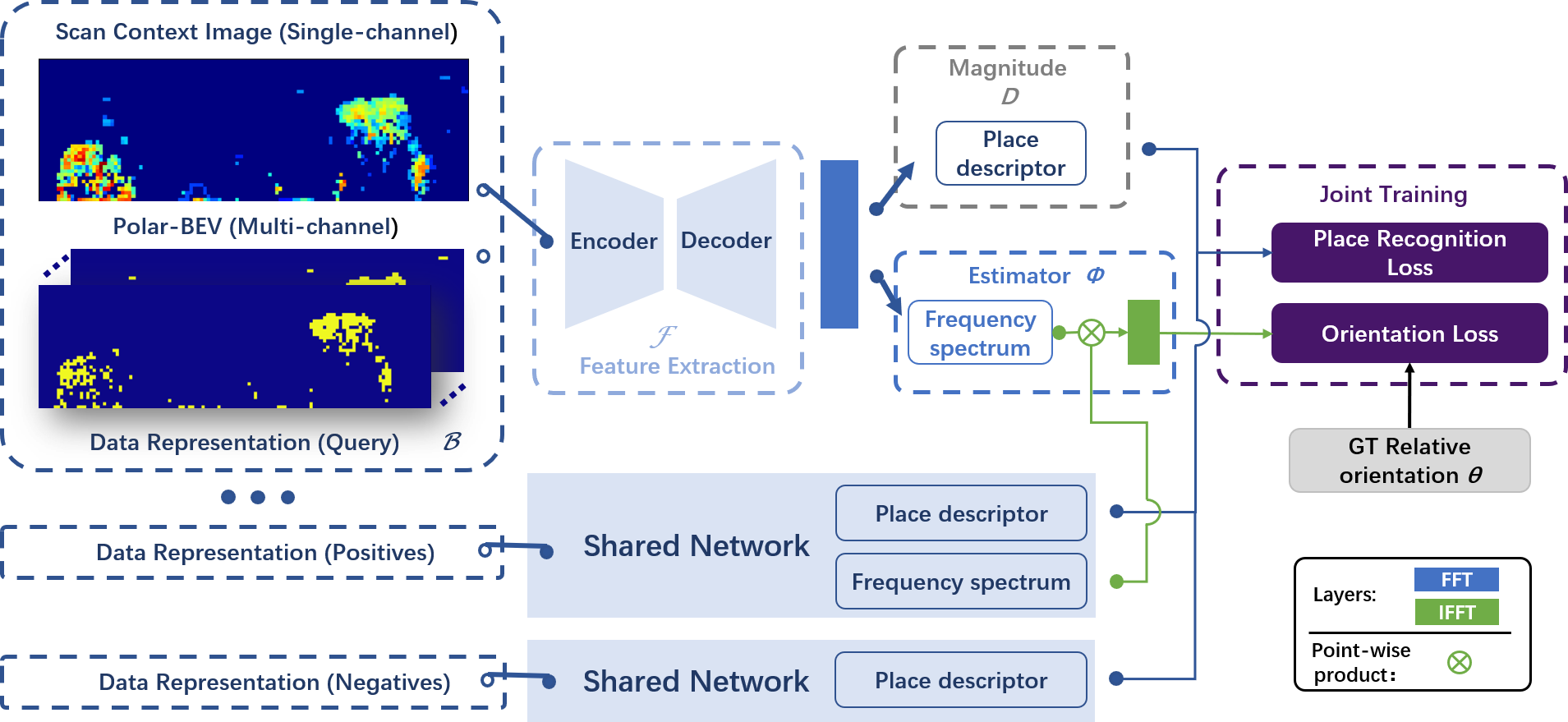}
\caption{\small{Overview of our proposed framework DiSCO. The data representation can be any polar BEV image.}}
\label{fig:Pipeline}
\vspace{-0.35cm}
\end{figure*}

\section{Methodology}

In this section, two main modules of our method are introduced in detail. The feature extraction module transforms point clouds into the frequency spectrum to generate place signature. The differentiable phase correlation module estimates the relative orientation between two scans using the correlation that also derived from the frequency spectrum. An overview of our system can be seen Fig. \ref{fig:Pipeline}.

\subsection{Problem Statement}

As the robot moves on the local planar ground surface, we consider the 3-DoF relative pose $(x,y,\theta)$ between a 3D LiDAR query scan $P_q$ and a map scan $P_s$ taken at a similar place. We define a similar place as the relative pose with slight translation $(x,y)$, but arbitrary $\theta$. For global localization, we have to deal with two problems. First, given $P_q$, recognize the $P_s \in M$ with arbitrary $\theta$, where $M$ is the map database. That is to achieve rotation invariance. Second, given $P_q$ and $P_s$, estimate the relative orientation $\theta$ without referring to any initial value. Generally, a scan registration is followed to achieve precise relative pose $(x,y,\theta)$. However, as the precise pose estimation is mature, we mainly focus on place recognition and orientation estimation.

\textbf{Rotation invariant signature learning:} Given the efficient downsampling ability of BEV representation and its robustness towards slight translation, we choose BEV as our data representation. Regarding the BEV representation as a function $\pazocal{B}$, we have $\pazocal{B}(P,\theta)$. When we change $\theta$, $\pazocal{B}$ changes accordingly, thus cannot achieve rotation invariance. Therefore, the core challenge of the first problem is to find a feature extractor $\pazocal{F}$, leading to
\begin{equation}
    \pazocal{F}(\pazocal{B}(P,0)) = \pazocal{F}(\pazocal{B}(P,\theta))
    \label{eq:BEV_notrans}
\end{equation}
Then we have
\begin{equation}
    \pazocal{F}(\pazocal{B}(P_q,0)) = \pazocal{F}(\pazocal{B}(P_s,0))
    \label{eq:BEVqs}
\end{equation}
which means the signatures of query scan $P_q$ and map scan $P_s$ becomes the same no matter how large relative orientation $\theta$ is, thus achieving rotation invariance.

\textbf{Relative orientation estimation:} On the contrary, when $P_s$ is recognized, we seek for rotation variance that can be utilized to estimate the orientation globally. Regarding this, we have to find another representation $\pazocal{H}$, and an estimator $\phi$ that
\begin{equation}
    \theta = \phi(\pazocal{H}(\pazocal{B}(P,0)),\pazocal{H}(\pazocal{B}(P,\theta)))
    \label{eq:BEVphi}
\end{equation}

An intuitive solution for this problem is that $\pazocal{H}$ is an identity mapping, and $\phi$ is an exhaustive solver. So the global optimal $\theta$ can be found by enumerating all possible rotations. However, for revisited places, point clouds are not exactly the same, so the identity mapping $\pazocal{H}$ will fail. Also, it's time-consuming for the exhaustive solver to enumerate all possibilities. So a solution is to find an efficient global solver $\phi$ as well as a matched mapping $\pazocal{H}$.

\subsection{Differentiable Signature in Frequency Domain}

\begin{figure}[t]
\centering
\includegraphics[width=0.47\textwidth]{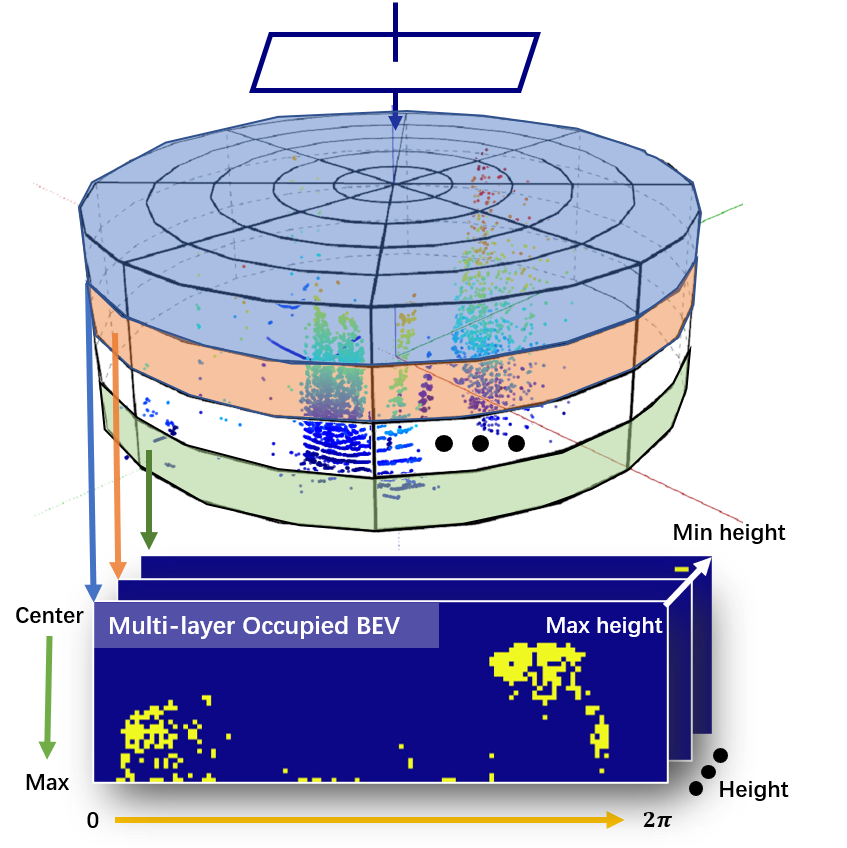}
\caption{\small{Demonstration of a BEV representation.}}
\label{fig:BEV}
\vspace{-0.35cm}
\end{figure}

We first propose several options for $\pazocal{B}$, an illustration is shown in Fig. \ref{fig:BEV}:
\begin{itemize}
  \item \textit{Multi-layer Occupied BEV} models occupied information. Voxelization is used to assign the original point cloud to 3D voxels. An empty voxel is regarded as unoccupied, otherwise, a voxel is occupied. Then we reorganize voxel values in the same height level into into image channels and construct a multi-layer BEV.
  \item \textit{Multi-layer Density BEV} models density feature. We voxelize point cloud into voxels as occupied BEV does and count points in every voxel. By using this method, we model the multi-layer density as well as the occupied information into our representation.
  \item \textit{Single-layer Height BEV (Scan Context)} models max-height feature. For each pillar in the voxel volume, we pick the height of the top point as the feature of this pilar, resulting in a 1-channel BEV image. This representation is introduced in \cite{Kim2018}, which is a special form of the BEV representations. We also evaluate this data representation in experiments.
\end{itemize}
In addition to these representations, any other BEV representation is qualified for our pipeline.

As we have constructed BEV representations, we need to find a feature extractor $\pazocal{F}$, an efficient global solver $\phi$ and a corresponding mapping $\pazocal{H}$. In this paper, we propose a network architecture containing all these modules.

\textbf{Network architecture:} To eliminate the effects of the rotation factor, we first convert the rotation factor into the translation factor by using polar transform $F_{pol}$, then followed by a convolution neural network (CNN), $F_{cnn}$, to learn the features in the polar domain. Finally, we apply Fourier Transformation $F_{fft}$ to convert the polar BEV image representation from its original domain to a representation in the frequency domain. Note that after operator $F_{pol}$, the rotation of the scan is transformed into translation in the polar domain, as shown in Fig. \ref{fig:BEV}. Then in the frequency domain, as the magnitude component of the frequency spectrum is translation-invariant, the signature actually becomes rotation invariant. Based on this character, we propose an operation for building signature $D$ as the output of $\pazocal{F}$ in Eq. (\ref{eq:BEVqs}) as
\begin{equation}
    D_\cdot = \pazocal{F}(\cdot) = |F_{fft} \circ F_{cnn} \circ F_{pol} (\cdot)|
    \label{eq:fsig}
\end{equation}

Note that in this pipeline, all operators are differentiable, and $F_{cnn}$ has parameters that can be learned.

\subsection{Metric Learning for Place Recognition}

Theoretically, Eq. (\ref{eq:fsig}) is rotation-invariant. However, as the translation cannot exactly equal to $0$, and the environmental structure may change slightly, exact rotation-invariance Eq. (\ref{eq:BEVqs}) is impossible. We instead set to build a weaker version
\begin{equation}\label{weakri}
  D_{P_s} = \arg\min_{P_i \in M} d(D_{P_q},D_{P_i})
\end{equation}
which means among all scans in the map database, the query scan $P_q$ and the scan at a similar place, $P_s$, have the minimal distance in the metric of $d$. As $D$ is rotation-invariant, we simply use Euclidean distance as the metric $d$, which can be implemented efficiently with KD-tree.

\textbf{Quadruplet training:} Inspired by Eq. (\ref{weakri}), we follow quadruplet loss $\pazocal{L}_{quad}$ in PointNetVLAD \cite{Uy2018} to train the parameters of $F_{cnn}$. In this paper, we use an encoder-decoder network architecture, U-Net, as $F_{cnn}$ to generate representation having the same resolution as the input. The loss is designed to force the network to learn a close distance between signatures taken at similar places and far apart distance between different ones. To eliminate the disturbance from high-frequency noise, we evaluate the distance only for the center part of $D$, which is the low-frequency component, indicating the general structure of the surrounding environment. The loss term is given as
\begin{equation}
    \begin{aligned}
        \pazocal{L}_{quad} =& \sum_{q,i}[\|D_{P_q}-D_{P_s}\|^2 - \|D_{P_q}-D_{P_{ni}}\|^2 +\alpha_1]_+ \\ &+ \sum_{q,i,j}[\|D_{P_q}-D_{P_s}\|^2 - \|D_{P_{ni}}-D_{P_{nj}}\|^2+\alpha_2]_+ \\
    \end{aligned}
\label{eq:quadruplet_loss}
\end{equation}
where $D_{P_q}$ and $D_{P_s}$ are the query scan and similar scan. $D_{P_{ni}}$ and $D_{P_{nj}}$ are the signatures of two negative samples i.e. scans taken at different places to that of query scan, $[\cdot]_+$ is the hinge loss of $\cdot$, $\alpha_1$ and $\alpha_2$ are the margins.

\subsection{Differentiable Relative Angle Estimation}

To estimate the relative orientation at the same time, we re-use a part of $\pazocal{F}$. Note that scans after applying $F_{cnn} \circ F_{pol}$, denoted as $G$, are represented in the polar domain, thus the rotation is parameterized by translation.
\begin{equation}\label{polar}
  G_s(\theta) = F_{cnn} \circ F_{pol} (\pazocal{B}(P_s,\theta))
\end{equation}

Estimating the relative orientation $\theta$ is equivalent to estimating relative translation between the two scans in the polar domain. As mentioned before, a global optimal solver for estimating the relative translation of two images is achieved by an exhaustive searcher.
\begin{equation}\label{ex}
  \theta = \arg \min_{\delta} \|G_q - G_s(\delta)\|
\end{equation}

Note that the sweeping variable is the translation, this exhaustive search is equivalent to cross-correlation when the metric in Eq. (\ref{ex}) is replaced with an inner product as
\begin{equation}\label{cc}
  \theta = \arg\max_{\delta} G_q * G_s(\delta)
\end{equation}
where $*$ is the convolution operator. The convolution of two images is equivalent to the point-wise multiplication of two frequency spectrums. Therefore, we can further share $F_{fft}$ with the signature mapping operation $\pazocal{F}$ as
\begin{equation}\label{dft}
  \theta = \arg\max_{\delta} F_{fft}^{-1} (F_{fft}(G_q) \otimes F_{fft}(G_s(\delta)))
\end{equation}
where $\otimes$ is the point-wise multiplication. With GPU based FFT, this operation can be very efficient without losing global optimality. Thus $\pazocal{H}$ in Eq. (\ref{eq:BEVphi}) is specified as
\begin{equation}
  \pazocal{H}(\cdot) = F_{fft} \circ F_{cnn} \circ F_{pol} (\cdot)
  \label{eq:fh}
\end{equation}

\begin{figure}[t]
\centering
\includegraphics[width=0.47\textwidth]{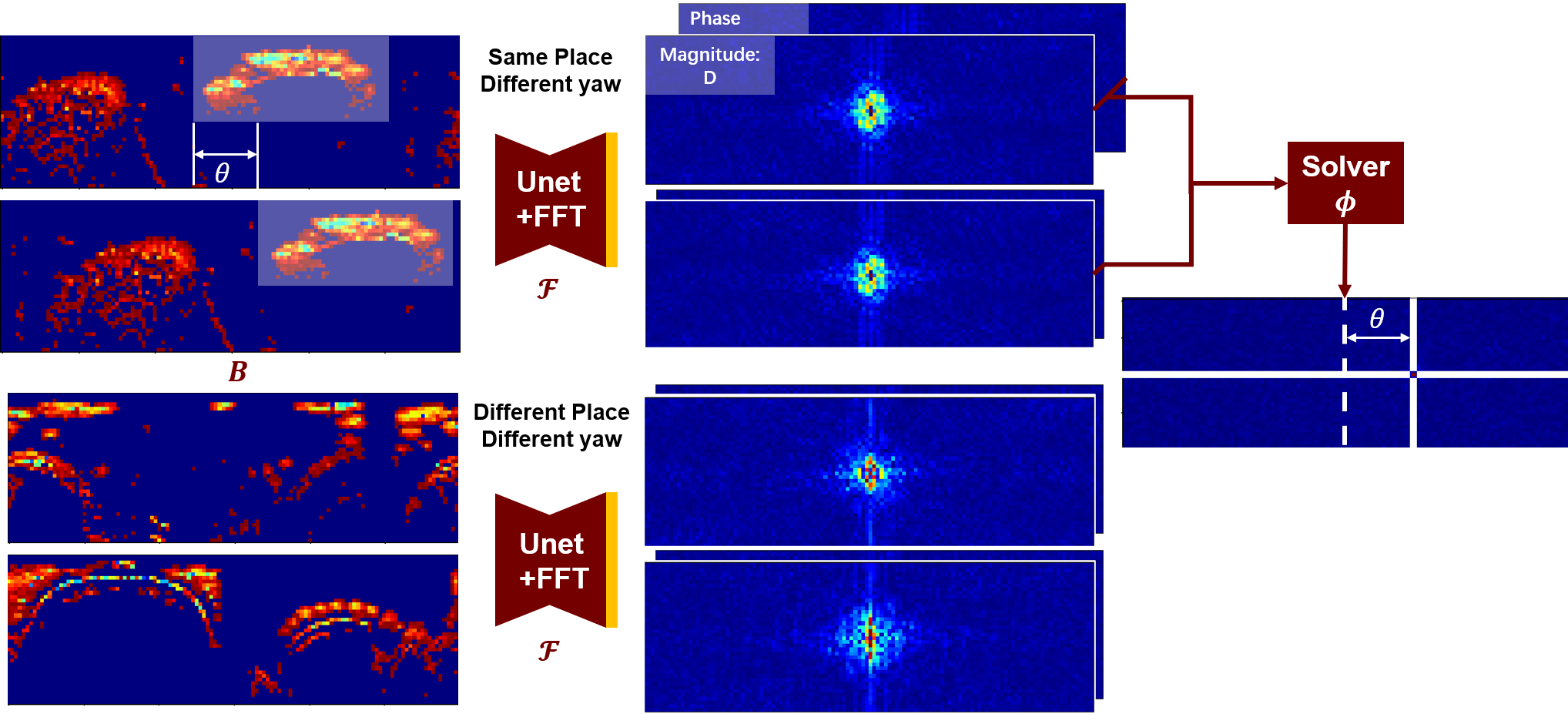}
\caption{\small{Output visualization from different step of our pipeline.}}
\label{fig:method_illustration}
\vspace{-0.45cm}
\end{figure}

The whole pipeline is called phase correlation and it is used for estimating the relative pose of images. The output visualization from different step is shown is Fig. \mbox{\ref{fig:method_illustration}}. This solver is efficient and global optimal but not differentiable due to $\arg\max$, which is an invalid form for $\phi$ in Eq. (\ref{eq:BEVphi}).

\textbf{Differentiable phase correlation:} To make the solver Eq. (\ref{dft}) differentiable, we replace the maximization based solver with an expectation based solver. Specifically, we regard the cross-correlation as a probability distribution $p(\delta)$ by applying a softmax block to keep every entry positive, and all entries summing up to $1$. After that, the estimation of $\theta$ is assigned with the expectation of $p(\delta)$.
\begin{equation}
\nonumber
p(\delta)= softmax(W \cdot F_{fft}^{-1} (F_{fft}(G_q) \otimes F_{fft}(G_s(\delta)))+b) \\
\end{equation}
\begin{equation}
\theta = \sum_{\delta} \delta p(\delta) \label{softmax}
\end{equation}
where $W$ and $b$ are parameters of softmax. Finally, we propose the form of $\phi$ as
\begin{equation}\label{est}
   \phi(\cdot,\cdot) = \sum_{\delta} \delta \cdot softmax(W \cdot F_{fft}^{-1}(\cdot \otimes \cdot)+b)
\end{equation}

For this task, we design the loss term as KL divergence between the probability distribution $p(\delta)$, with the supervision of a one-hot distribution $\textbf{1}(\delta-\theta^*)$ peaking at the ground truth angle $\theta^*$
\begin{equation}\label{angleloss}
  \pazocal{L}_{rot} = KLD(p(\delta),\textbf{1}(\delta-\theta^*))
\end{equation}
\vspace{-0.3cm}

\subsection{Implementation and Training}

Based on the operators Eq. (\ref{eq:fsig}), (\ref{eq:fh}) and (\ref{est}), we have a fully sharable network backbone without any extra parameters to satisfy the both tasks, which is the benefit of the strong interpretable structure. In this pipeline, the convolutional network is enforced to learn features from the polar domain. Note that instead of adding $F_{cnn}$ before the polar transform, we add $F_{cnn}$ after the polar transform. We consider that convolution is effective in dealing with translation variants but fails in rotation variation in the image.

We train the network in joint learning manner. Two losses Eq. (\ref{eq:quadruplet_loss}) and (\ref{angleloss}) are combined into a joined loss $\pazocal{L}$:
\begin{equation}
\pazocal{L} = \pazocal{L}_{quad} + \lambda \pazocal{L}_{rot}
\label{eq:total_loss}
\end{equation}
where $\lambda$ is a hyper-parameter to balance the two terms.

We use ADAM \cite{Kingma2015} as our optimizer and set the learning rate to 0.00001. The margins for quadruplet loss are $\alpha_1 = 0.5$, $\alpha_2 = 0.2$. We randomly rotate the input LiDAR scans using a method like N-way augmentation on BEVs as in \mbox{\cite{Kim2019}} in both training and evaluation steps. All modules of DiSCO are trained together with these settings.

\section{Experiments}

The experiments we conduct is designed to support our claims about DiSCO: (i) It can simultaneously predict place candidates and orientation. (ii) It is efficient and lightweight. Furthermore, we separately analyze the place recognition and orientation estimation module and provide a detailed ablation study on dimensionality and resolution.

\begin{table}[t]
\centering
\caption{\small{Comparison of different methods for evaluation.}}
\begin{tabular}{lccccc}
\toprule
\textbf{Approach} & \textbf{Place retrieval} & \textbf{Orientation} \\
\midrule
PointNetVLAD \cite{Uy2018}  & \checkmark & $\times$  \\
ScanContext \cite{Kim2019}  & \checkmark & $\times$  \\
OverlapNet \cite{Chen2020} & Limited & \checkmark   \\
OREOS \cite{Schaupp2019} &\checkmark & \checkmark  \\
DiSCO (our) & \checkmark & \checkmark  \\
\bottomrule
\end{tabular}
\label{tb:comparison}
\vspace{-0.35cm}
\end{table}

\subsection{Dataset and Experimental Settings}

\textbf{Oxford Dataset:} We use the benchmark dataset proposed by PointNetVLAD. The original point cloud data is from open-source Oxford RobotCar \cite{maddern20171}. It contains LiDAR sensor data collected at different times of a 10km region. In \cite{Uy2018}, points in ground planes are removed and each submap is at fixed regular intervals of 20m for testing and 10m for training. All submaps are downsampled to a fixed 4096 points. The ground truth with the UTM coordinate is also attached to each submap. To get the correct annotation of data, point clouds are split into positives and negatives pairs. Those point clouds that are at most 10m apart is defined as positive pairs while negative pairs are at least 50m apart. For evaluation, the query submap is successfully localized if the retrieved point cloud is in 25m. Following the experimental settings in \cite{Uy2018}, we use 44 sets of runs to train our network. There are 21711 submaps for training and 3030 for testing.

\begin{figure*}
\centering
\includegraphics[width=1.0\textwidth]{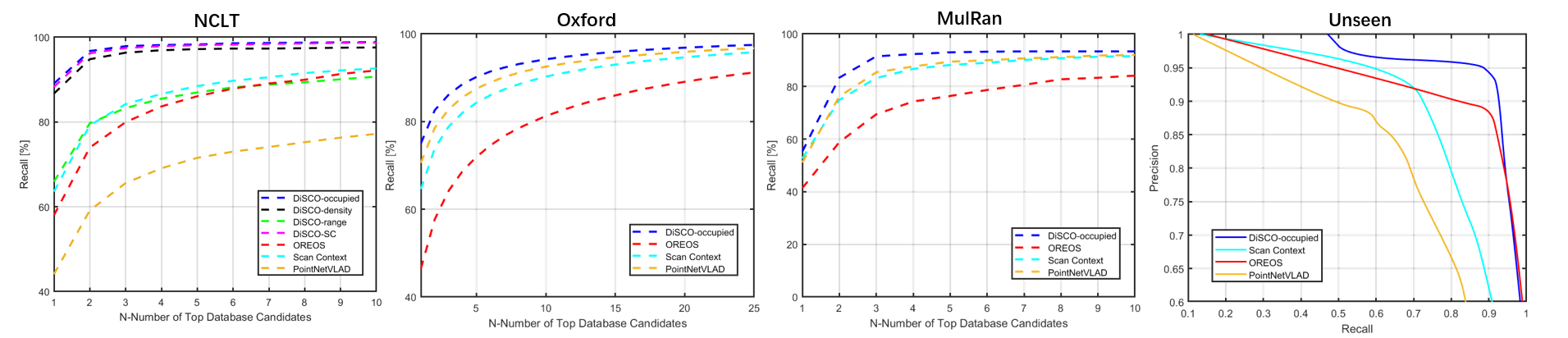}
\caption{\small{Place recognition performance on NCLT, Oxford and MulRan datasets using different variants of DiSCO and comparison methods and unseen place classification precision-recall curves.}}
\label{fig:PR_Curve}
\vspace{-0.45cm}
\end{figure*}


\textbf{NCLT Dataset:} NCLT is the abbreviation of The University of Michigan North Campus Long-Term Vision and LIDAR Dataset \cite{carlevaris2016university} which consists of 27 sets of driving data on the University campus over 14 months. A Velodyne HDL-32 sensor is mounted on the driving platform to provide point clouds at 10 Hz.

Our process steps for metric learning follow the settings in \cite{Schaupp2019} which requires sampling of positive pairs and negative pairs. In the NCLT dataset, point clouds are defined as positive pairs as they are 1.5m apart and negative pairs as they are at least 2m apart. After 5 epoch of training, we adopt a hard-negative mining strategy to boost the performance of the network. The input of our network is the original point cloud with the ground plane removed by an area filter. We restrict point to be within 80m far and 20m height. We use only 1 run (2012-02-04) to train our model and 6 runs (2012-03-17, 2012-05-26, 2012-06-15, 2012-08-20, 2012-09-28, 2012-10-28) to evaluate. The query scan is successfully localized if the retrieved point cloud is in {\color{red}1.5m}.

\textbf{MulRan Dataset:} MulRan dataset is a multimodal range dataset for urban place recognition \mbox{\cite{kim2020mulran}} which consists of month-level challenging driving data in an urban environment. An Ouster OS1-64 is mounted on the car for 3D perception at 10 Hz. 

Following the data preprocessing of the NCLT dataset, we choose DCC01 to train our model and DCC02, DCC03 to evaluate. Unlike the NCLT dataset which is collected by a lightweight wheeled robot, the MulRan dataset is acquired by a fast-moving car causing relatively large intervals between consecutive point clouds. We set the positive pair threshold to 3m and the rest of pairs is considered as negatives. The query scan is successfully localized if the retrieved point cloud is in 3m.

\begin{table}[t]
\centering
\caption{\small{Detailed experiment results.}}
\resizebox{0.48\textwidth}{28mm}{
\begin{tabular}{clccc}
\toprule
\textbf{Dataset} & \textbf{Approach} & \textbf{Recall@1} & \textbf{Recall@1\%} & \textbf{AUC}\\
\midrule
\multirow{6}{*}{NCLT} & \textbf{DiSCO-occupied (our)} & \textbf{89.08} & \textbf{99.14} & \textbf{92.91} \\
& DiSCO-density (our) & 86.71 & 97.87 & 90.76 \\
& DiSCO-range (our)& 65.93 & 90.38 & 72.82 \\
& DiSCO-SC (our) & 88.28 & 99.02 & 91.02 \\
& OREOS \cite{Schaupp2019} & 58.02 & 91.44 & 66.00\\
& Scan Context \cite{Kim2019} & 63.58 & 92.35 & 71.39\\
& PointNetVLAD \cite{Uy2018} & 44.1 & 76.27 & 51.62 \\
\hline
\multirow{3}{*}{Oxford} & \textbf{DiSCO-occupied (our)}& \textbf{75.01} & \textbf{88.44} & \textbf{78.81}\\
& OREOS \cite{Schaupp2019} & 46.46 & 68.47 & 52.10\\
& Scan Context \cite{Kim2019} & 64.59 & 81.88 & 69.30\\
& PointNetVLAD \cite{Uy2018}  & 70.57 & 85.42 & 74.61\\
\hline
\multirow{3}{*}{MulRan} & \textbf{DiSCO-occupied (our)}& \textbf{55.28} & \textbf{93.34} & \textbf{69.32}\\
& OREOS \cite{Schaupp2019} & 41.39 & 84.89 & 50.11\\
& Scan Context \cite{Kim2019} & 52.63 & 92.06 & 63.76\\
& PointNetVLAD \cite{Uy2018}  & 51.10 & 92.00 & 63.65\\
\bottomrule
\end{tabular}}
\label{tb:PR_detail}
\vspace{-0.35cm}
\end{table}

\subsection{Comparison Methods}

We compare our metric learning algorithm with four competitive methods in recent years: PointNetVLAD \cite{Uy2018}, Scan Context \cite{Kim2019}, OREOS \cite{Schaupp2019} and OverlapNet \cite{Chen2020}. General characters of these approaches are summarized in Tab. \ref{tb:comparison}.
\begin{itemize}
\item \textbf{PointNetVLAD} is a combination of PointNet and NetVLAD, so it can directly learn from point coordinates. We follow the original paper to setup our evaluation experiments and use the open-source code in PyTorch version to get 1024-dimensional descriptors.

\item \textbf{Scan Context} is a handcrafted descriptor extracted from input point clouds. The original Scan Context regards the place recognition task as a combination of a place classification and a new place detection problem so that they use classification networks to predict place index in the database and new place. To fit into our metric learning structure and evaluate performance, we reimplement the Scan Context as one kind of data representation of our system. We evaluate the original Scan Context and also use the Scan Context representation as our single-layer height BEV in our network named DiSCO-SC. The ring and sector parameter is set to 40, 120.

\item \textbf{OREOS} is implemented by us following the paper's elaboration. Unlike the original method that adopts ICP as post process, we only evaluate orientation estimation that the network predicts. We also implement a version of our method for evaluation using range image as input named DiSCO-range. Output dimension is 1024.

\item \textbf{OverlapNet} estimates the overlap between two scans by learning the features in range images. Using a correlation head, OverlapNet can also estimate yaw angles between two range images. The overlap information can be further used in place recognition tasks, but due to the exhaustive mechanism in the retrieval task, we only evaluate OverlapNet with its yaw estimation part.

\item \textbf{Ours} We evaluate some variants of DiSCO. We construct BEVs with ring and sector same as Scan Context (40, 120) and discretizes the height 20m with 20 layers. So the size of Multi-layer BEV is 40$\times$120$\times$20. Output dimension is set to 1024.

\end{itemize}

\begin{figure}[t]
\centering
\includegraphics[width=0.46\textwidth]{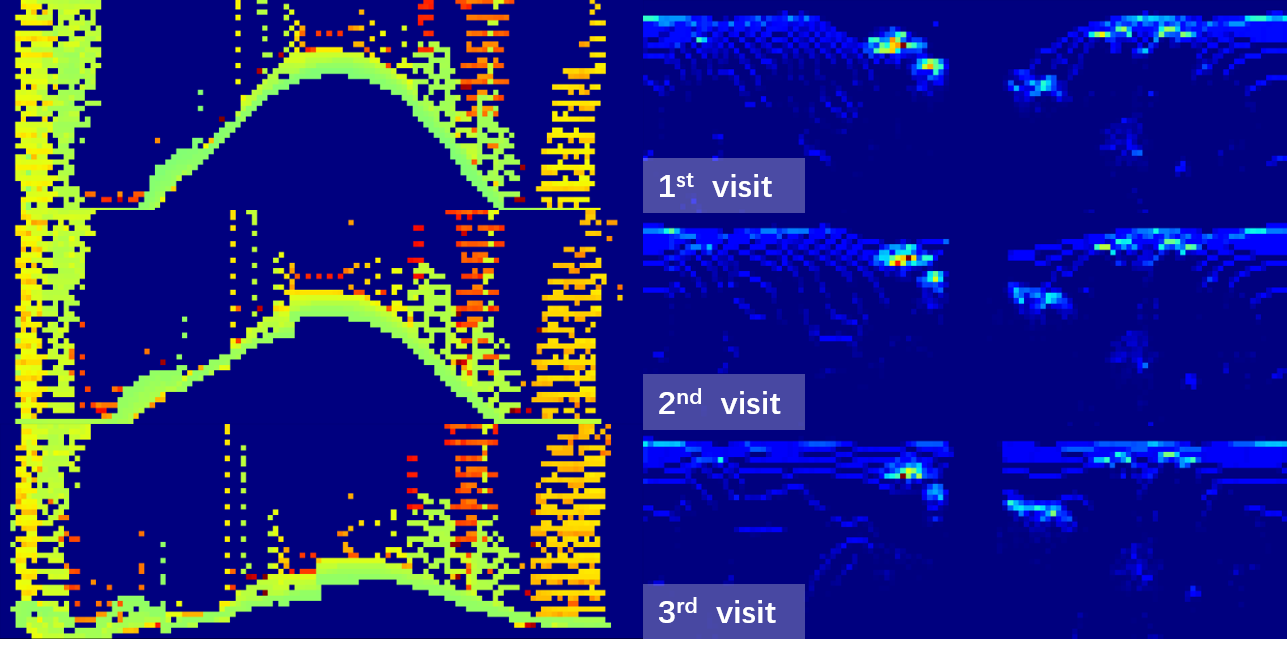}
\caption{\small{Demonstration of the harmful pitch influence on range image. Three range images in the left column are at the same location. The corresponding BEV images shown in the right column are not badly affected. For Visualization, we use the same size for both kinds of representation.}}
\label{fig:range_image}
\vspace{-0.45cm}
\end{figure}

\subsection{Place Recognition Analysis}

In a practical place retrieval scenario, we often test more than one place candidate to ensure the robustness. To this end, we evaluate the performance of the place recognition module with top $k$ candidates. Corresponding to the previous comparison methods, the $k$ in the Oxford dataset is 25, and 10 in the NCLT dataset and MulRan dataset \cite{Uy2018,Schaupp2019}.

We evaluate the place recognition performance of baseline methods and our DiSCO variants using the precision curve for the NCLT, MulRan and Oxford datasets. Unseen place rejection stated in \mbox{\cite{Kim2019, cummins2008fab}} is also important to prevent false data associations. To evaluate this we modify the original Scan Context and follow the settings in \mbox{\cite{Kim2019}}. Using 2012-02-04 as the database, run 2012-05-26 which contains 170 new places as online loop detection. Experiment results are shown in Fig. \mbox{\ref{fig:PR_Curve}} and Tab. \mbox{\ref{tb:PR_detail}}. We only show the DiSCO-occupied results on Oxford and MulRan dataset because of its relatively stable performance. Full results can be found in the Appendix. Note that, even if density information is utilized, the precision is limited. We believe that density information can be easily disturbed which results in poorer performance than occupied information.In the Oxford dataset, due to the large sampling interval between two places, our polar BEV representation suffers from the relatively small overlap area between some scans and can not achieve its best ability. This may be regarded as a limitation, but we consider that it can be an option determined by the application. In the MulRan dataset, the top 1 precision are not satisfied, we attribute it to the possible error imported by aligning the timestamp.

In NCLT dataset, the robot car suffers from large pitch disturbance which is harmful to the range image, the demonstration of the effect is shown in Fig. \ref{fig:range_image}. This effect is not significant in the BEV images because of the error-tolerant sampling method of polar transform, and this also shows the robustness of the polar BEV representations. A specific explanation can be found in the Appendix.  The contrast between the original Scan Context and Scan Context with Fourier transform (DiSCO-SC) indicates the efficiency and effectiveness of our place recognition module. The drastic reduction of PointNetVLAD \cite{Uy2018} on NCLT mainly results from different types of LiDAR. Unlike the submaps built by a 2D LiDAR sensor like SICK, point clouds collected by a 3D LiDAR sensor used in the NCLT Dataset is much sparser.

\begin{table}[t]\small
\centering
\caption{\small{Absolute orientation estimation errors with mean and standard deviation in degrees.}}
\begin{tabular}{lllll}
\toprule
\textbf{Dataset}  & \textbf{Approach}  & \textbf{Mean [deg]} & \textbf{Std [deg]} \\
\midrule
\multirow{5}{*}{NCLT} & OREOS \cite{Schaupp2019} & 15.95 & 21.31 \\
& OverlapNet \cite{Chen2020}  & 11.59 & 24.10 \\
& \textbf{DiSCO-occupied (our)} & \textbf{2.81} & \textbf{4.01}  \\
& DiSCO-density (our) & 3.86 & 8.31  \\
& DiSCO-SC (our) & 5.03 & 6.58 \\
\hline
\multirow{3}{*}{MulRan} & OREOS \cite{Schaupp2019} & 10.92 & 9.46 \\
& OverlapNet \cite{Chen2020}  & 2.74 & 8.57 \\
& \textbf{DiSCO-occupied (our)} & \textbf{1.99} & \textbf{5.11}  \\
\bottomrule
\end{tabular}
\label{tb:yaw_err}
\vspace{-0.35cm}
\end{table}

\begin{table}[t]\small
\centering
\caption{\small{Voxelization resolution analysis on NCLT.}}
\resizebox{0.48\textwidth}{7mm}{
\begin{tabular}{ccccc}
\toprule
\textbf{Rotation Res} & \textbf{Recall @1} & \textbf{Mean [deg]} & \textbf{Std [deg]} & \textbf{Runtime [ms]}\\
\midrule
3 deg/pix & 89.08 & 2.81 & 4.01 & \textbf{9.5} \\
\textbf{1 deg/pix} & \textbf{89.63} & \textbf{2.32} & \textbf{3.59} & 10.2\\
\bottomrule
\end{tabular}}
\label{tb:res_analysis}
\vspace{-0.35cm}
\end{table}

\subsection{Orientation Estimation Analysis}

To evaluate the accuracy of the orientation estimation module, we use the NCLT dataset with ground-truth orientations attached. Similar to the evaluation indicator in OREOS \cite{Schaupp2019}, we assess the mean and standard deviations of the estimation errors. As OverlapNet \mbox{\cite{Chen2020}} can also estimate relative yaw between two scans, we evaluate its orientation performance on the retrieved scans by DiSCO using depth and normals as input. The experiment results are listed in Tab. \ref{tb:yaw_err}. Note that the results of orientation estimation are those produced by the author of OREOS.

Our correlation-based orientation estimator not only provides interpretability but also improves the orientation estimation performance. The superior performance of DiSCO supports the idea that our differentiable geometric solver is more appropriate than the network-based regression solver. The evaluation of OverlapNet also proves this idea.

In addition, the orientation estimation accuracy is sensitive to the resolution of the input. The polar transform discretizes the rotation space into a fixed-size so that the signature has its theoretical maximal precision. In our settings the theoretical maximal precision is $360\degree/120pix = 3\degree/pix$. In the ablation section, we further discuss the influence of the resolution.


\subsection{Ablation Study}


\textbf{Voxel resolution analysis:} The resolution of voxelization is intuitively an essential factor of the final prediction. We test several resolutions in rotation to show the impact. The experiment results are shown in Tab. \ref{tb:res_analysis}. The dense voxelization in rotation can bring higher precision in orientation estimation, but the larger input also consumes more time.

\begin{table}[t]\small
\centering
\caption{\small{Cutoff frequency analysis with efficiency evaluation. LPF represents low-pass filter and HPF represents high-pass filter.}}
\begin{tabular}{cccc}
\toprule
\textbf{Dimension} & \textbf{Recall @1} & \textbf{Recall @1\%} & \textbf{Runtime [ms]} \\
\midrule
LPF-256 & 84.33 & 97.33 & \textbf{9.0} \\
\textbf{LPF-1024} & \textbf{86.71} & \textbf{97.87} & 9.5 \\
LPF-4800 & 78.20 & 95.73 & 12.9 \\
HPF-1024 & 2.36 & 7.73  & 9.5 \\
\bottomrule
\end{tabular}
\label{tb:dim_analysis}
\vspace{-0.35cm}
\end{table}

\begin{figure}[t]
\centering
\includegraphics[width=0.45\textwidth]{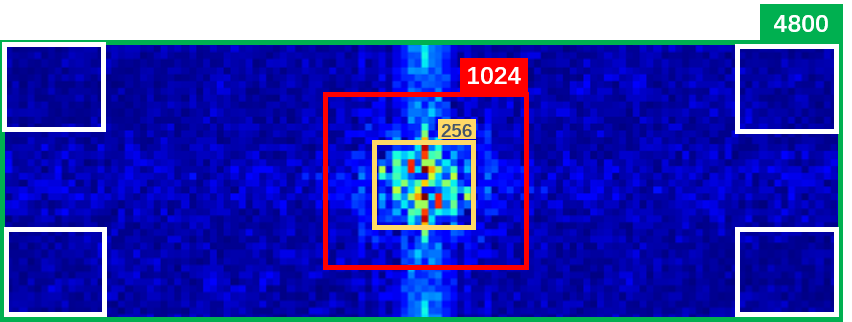}
\caption{\small{Visualization of signature and ablation settings. Four white boxes are concatenated to be a high-frequency signature.}}
\label{fig:descriptor}
\vspace{-0.40cm}
\end{figure}

\textbf{Cutoff frequency analysis:} We visualize the signatures of scans as shown in Fig. \ref{fig:descriptor} and find that almost all information is gathered at the center of the magnitude. The actual size of this signature is the same as the input, but the informative area is relatively small. So we implement some low-pass and high-pass filters with different cutoff frequencies for our network and study the discriminative ability and efficiency of our method. The results are shown in Tab. \ref{tb:dim_analysis}.

The middle part of the signature is the low-frequency component which indicates the general structure of the environment while the rest is often high-frequency details or noise. As shown in experiments, our network mainly learns to recognize the general structure of the environment.




\subsection{Runtime evaluation}

We test our method on a system equipped with an AMD Ryzen 7 3700X CPU with 3.6GHz and an Nvidia GeForce RTX 2060 Super with 8GB memory. A detailed runtime analysis is shown in Tab. \ref{tb:runtime}. In the Oxford Dataset, the scans in benchmark dataset have already been processed, so the preprocess step is fast. As for the NCLT and MulRan dataset, the input is the original scans containing almost 30k points. Implemented in parallel, we achieve almost the same time as in the Oxford dataset. By using the KD-tree structure, retrieval through a submap database takes logarithm time complexity. Therefore, our system can realize online place recognition and orientation estimation in long-term tasks.

\begin{table}[t]
\caption{\small{Average computational execution times of our DiSCO in the different dataset in [ms].}}
\centering
\begin{tabular}{ccccc}
\toprule
\centering
\textbf{Dataset} & \textbf{Preprocess} & \textbf{Generate} & \textbf{Estimate} & \textbf{Total}\\
& & \textbf{signature} & \textbf{orientation} & \\
\midrule
NCLT & 0.80 & \multirow{3}{*}{6.20} & \multirow{3}{*}{2.50} & 9.50 \\
MulRan & 0.80 & & & 9.50 \\
Oxford & 0.40 & & & 9.10 \\
\bottomrule
\end{tabular}
\label{tb:runtime}
\vspace{-0.40cm}
\end{table}

\section{Conclusion}

In this paper, we propose a novel framework for simultaneously estimating the relative orientation and place candidates in challenging long-term conditions. The main idea of the framework is to convert the rotation-invariant signature to the translation-invariant frequency spectrum. Based on such structure design as constraints, the two tasks can share the whole network without additional parameters, achieving effective and real-time performance.

In the future, we plan to combine the metric pose estimation into our method to achieve a long-term SLAM system.




\bibliographystyle{IEEEtran}
\bibliography{IEEEabrv,IEEE}

\clearpage
\section*{APPENDIX}
\subsection{Dataset Preparation}
\begin{figure}[h]
\centering
\includegraphics[width=0.47\textwidth]{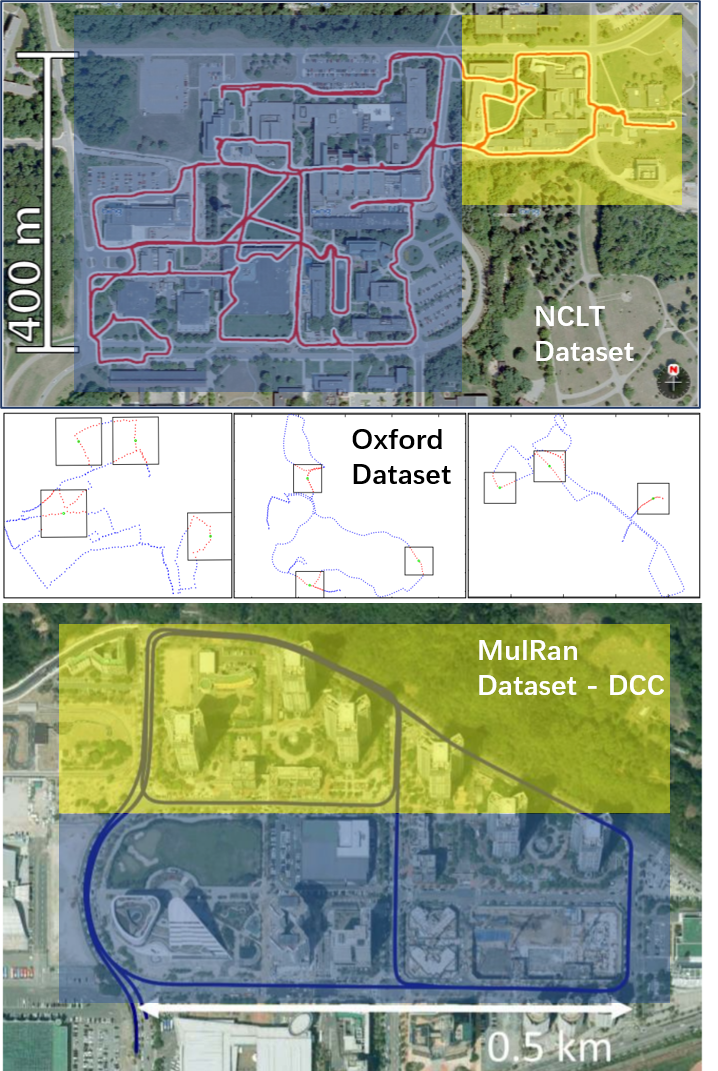}
\caption{NCLT data split: the query places in evaluation runs are from the yellow part and places used for training is from the blue part.}
\label{fig:data_split}
\end{figure}

All dataset used in our experiments is split into training and evaluating parts shown in Fig. \ref{fig:data_split}. The query places in evaluation runs are from the yellow areas and the rest of the dataset places are for training in only 1 run.

The preprocessing step in NCLT and Oxford dataset is following the settings in \cite{Schaupp2019,Uy2018}. Same as in these datasets, we first sample the MulRan dataset with 1m to construct training sets. The scans that fall within 3m are considered as positive pairs while others are regarded as negative pairs. As for the evaluation, we sample the places in the evaluation area at 5m intervals. Places retrieved within 3m are considered as successful. The runs used in our experiment is from the DCC area. DCC01 is for training and others are used for evaluation.

\begin{figure*}
  \centering
  \begin{subfigure}[b]{0.48\textwidth}
      \centering
      \includegraphics[width=\linewidth]{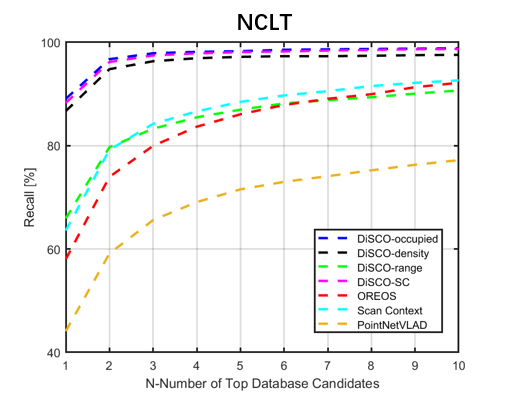}
      \label{fig:descriptor}
      \caption{Place recognition performance on NCLT dataset.}
  \end{subfigure}
  \hfill
  \begin{subfigure}[b]{0.48\textwidth}
      \centering
      \includegraphics[width=\linewidth]{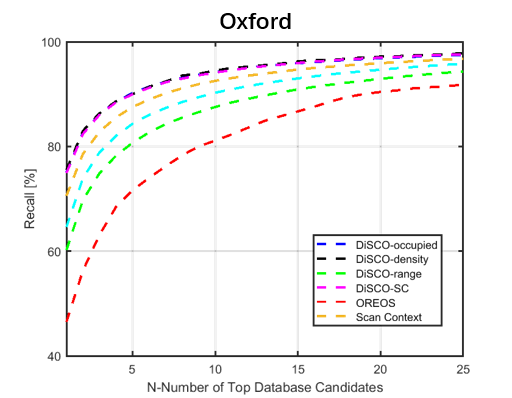}
      \label{fig:descriptor_center}
      \caption{Place recognition performance on Oxford dataset.}
  \end{subfigure}
  \hfill
  \begin{subfigure}[b]{0.48\textwidth}
      \centering
      \includegraphics[width=\linewidth]{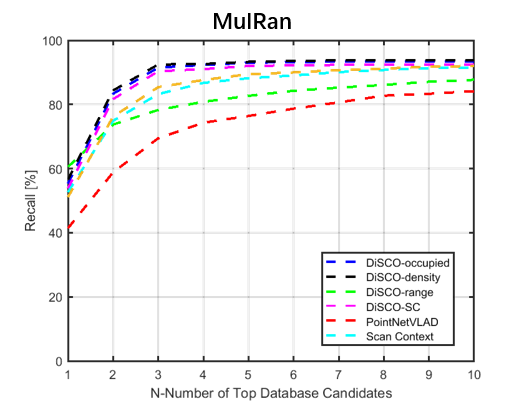}
      \label{fig:descriptor_center}
      \caption{Place recognition performance on MulRan dataset.}
  \end{subfigure}
  \caption{Precision curves of place recognition performance.}
  \label{fig:full_results}
\end{figure*}

\subsection{Experiment results}

We only choose occupied DiSCO to show in the Fig. \ref{fig:PR_Curve} mainly for two reasons. First, the space of the paper is limited and we put our focus on comparison with other methods. Second, the performance of occupied DiSCO is relatively stable which can be found in the full results shown in Fig. \ref{fig:full_results} and Tab. \ref{tb:Full_PR_detail}.

In the MulRan dataset, the top 1 precision are not satisfied, we attribute it to the possible error imported by aligning the timestamp.

\begin{table}[hp]
\centering
\caption{\small{Detailed experiment results.}}
\begin{tabular}{clccc}
\toprule
\textbf{Dataset} & \textbf{Approach} & \textbf{Recall@1} & \textbf{Recall@1\%} & \textbf{AUC}\\
\midrule
\multirow{6}{*}{NCLT} & \textbf{DiSCO-occupied (our)} & \textbf{89.08} & \textbf{99.14} & \textbf{92.91} \\
& DiSCO-density (our) & 86.71 & 97.87 & 90.76 \\
& DiSCO-range (our)& 65.93 & 90.38 & 72.82 \\
& DiSCO-SC (our) & 88.28 & 99.02 & 91.02 \\
& OREOS \cite{Schaupp2019} & 58.02 & 91.44 & 66.00\\
& Scan Context \cite{Kim2019} & 63.58 & 92.35 & 71.39\\
& PointNetVLAD \cite{Uy2018} & 44.1 & 76.27 & 51.62 \\
\hline
\multirow{6}{*}{Oxford} & DiSCO-occupied (our)& 75.01 & 88.44 & 78.81\\
& \textbf{DiSCO-density (our)} & \textbf{75.52} & \textbf{88.62} & \textbf{79.25} \\
& DiSCO-range (our)& 60.22 & 80.70 & 65.00 \\
& DiSCO-SC (our) & 74.92 & 88.41 & 78.65 \\
& OREOS \cite{Schaupp2019} & 46.46 & 68.47 & 52.10\\
& Scan Context \cite{Kim2019} & 64.59 & 81.88 & 69.30\\
& PointNetVLAD \cite{Uy2018}  & 70.57 & 85.42 & 74.61\\
\hline
\multirow{6}{*}{MulRan} & DiSCO-occupied (our)& 55.28 & 93.34 & 69.32\\
& \textbf{DiSCO-density (our)} & 56.58 & \textbf{93.53} & \textbf{70.74} \\
& DiSCO-range (our)& \textbf{60.52} & 87.68 & 67.13 \\
& DiSCO-SC (our) & 53.78 & 92.32 & 67.72 \\
& OREOS \cite{Schaupp2019} & 41.39 & 84.89 & 50.11\\
& Scan Context \cite{Kim2019} & 52.63 & 92.06 & 63.76\\
& PointNetVLAD \cite{Uy2018}  & 51.10 & 92.00 & 63.65\\
\bottomrule
\end{tabular}
\label{tb:Full_PR_detail}
\end{table}

\subsection{Pitch Problem}

\begin{figure*}[htp]
\centering
\includegraphics[width=1.0\textwidth]{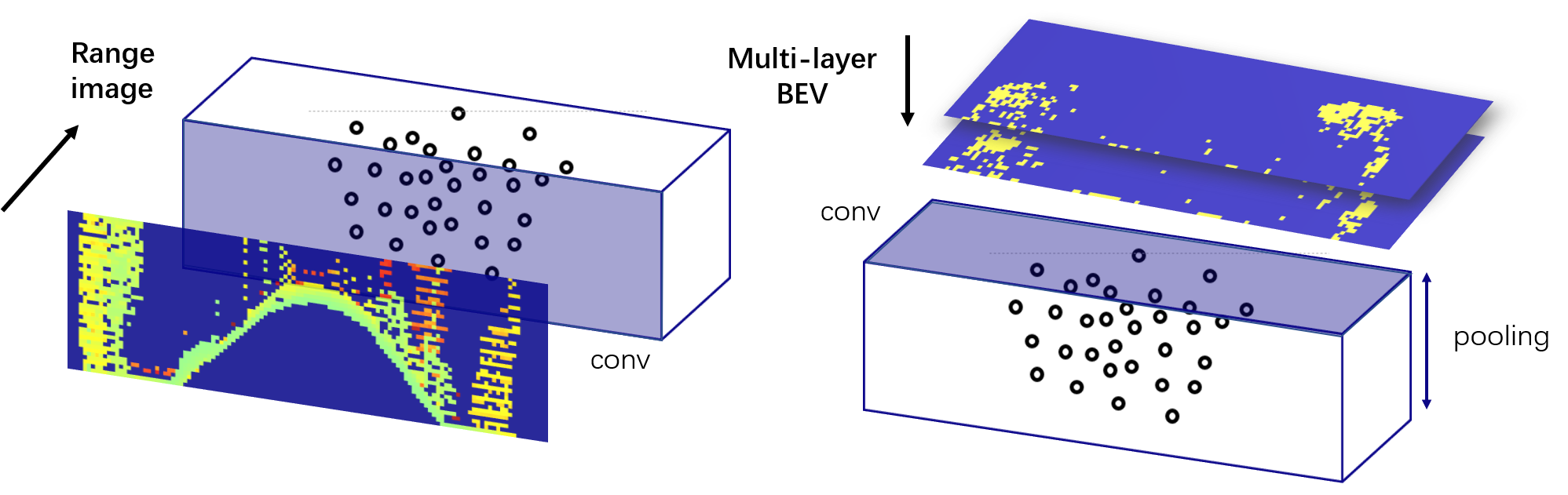}
\caption{\small{Illustration of how the pitch influence different representation.}}
\label{fig:range_explain}
\end{figure*}

The multi-layer BEVs alleviate the influence brought by pitch disturbance mainly by its viewpoint. As shown in schematic Fig. \ref{fig:range_explain}, the range image is constructed at the front camera view, so that the same obstacle pixels will move in the y dimension of the range image due to the pitch disturbance. The change in the y dimension will be harmful to the feature extraction network as convolution has limited translation invariance. The Multi-layer BEV is constructed at the bird's view as shown in the schematic Fig. \ref{fig:range_explain}. Points from the close height level form a channel of the multi-layer BEV image so that the same obstacle points will change in the channel dimension due to the pitch disturbance. Unlike the range image, the convolution on the BEV images is operated at each BEV image and a pooling operation is adopted at the channel dimension. The pooling operation is translation invariance thus achieving a certain extent pitch invariance. As for ground plane extraction, we just use a simple height filter to remove the ground. 

\end{document}